\documentclass[conference]{IEEEtran}
\IEEEoverridecommandlockouts
\usepackage{cite}
\usepackage[strings]{underscore}
\usepackage{csquotes}
\usepackage{amsmath,amssymb,amsfonts}
\usepackage{algorithm}
\usepackage{algorithmicx}
\usepackage{algpseudocode}
\usepackage{graphicx}
\usepackage{textcomp}
\usepackage{xcolor}
\usepackage{caption}
\usepackage{subcaption}

\def\BibTeX{{\rm B\kern-.05em{\sc i\kern-.025em b}\kern-.08em
    T\kern-.1667em\lower.7ex\hbox{E}\kern-.125emX}}
\begin{document}

\title{ Unsupervised anomalies detection in IIoT edge devices’ networks using federated learning.}

\author{\IEEEauthorblockN{\textsuperscript{} Niyomukiza Thamar}
\IEEEauthorblockA{\textit{Department of Computer Science and Engineering} \\
\textit{The American University in Cairo}\\
Cairo, Egypt \\
 }
\and
\IEEEauthorblockN{\textsuperscript{} Dr. Hossam Samy Elsaid Sharara}
\IEEEauthorblockA{\textit{Department of Computer Science and Engineering} \\
\textit{The American University in Cairo}\\
Cairo, Egypt \\}
}

\maketitle

\begin{abstract}
 IoT/ IIoT devices' data may be considered highly private; as such, training machine learning models to help detect intrusions in the network of those devices requires so much attention. In a connection of many IoT devices that each collect data, normally training a machine learning model would involve transmitting the data to a central server which requires strict privacy rules. However, some owners are reluctant of availing their data out of the company due to data security concerns.

Federated learning(FL) as a distributed machine learning approach performs training of a machine learning model on the device that gathered the data itself. In this scenario, data is not share over the network for training purpose.  Fedavg as one of FL algorithms permits a model to be copied to participating devices during a training session. The devices could be chosen at random, and a device can be aborted.  The resulting models are sent to the coordinating server and then average models from the devices that finished training. The process is repeated until a desired model accuracy is achieved. By doing this, FL approach solves the privacy problem for IoT/ IIoT devices that held sensitive data for the owners. In this paper, we leverage the benefits of FL and implemented Fedavg algorithm on a recent dataset that represent the modern IoT/ IIoT device networks. The results were almost the same as the centralized machine learning approach. We also evaluated some shortcomings of Fedavg such as unfairness that happens during the training when struggling devices do not participate for every stage of training. This inefficient training  of local or global model could lead in a high number of false alarms in intrusion detection systems for IoT/IIoT gadgets developed using Fedavg. Hence, after evaluating the FedAv deep auto encoder with centralized deep auto encoder ML, we further proposed and designed a Fair Fedavg algorithm that will be evaluated in the future work.
\end{abstract}


\section{Introduction}
\label{ch:into} 

IoT edge devices have found their ways into every corner of humans’ lives. They are very popular and have become essential in our daily lives both individually and on the industrial level \cite{b1}. Despite their wide adoption in many sectors such as healthcare for smart healthcare surveillance of patients , smart homes, educational and industrial environments, they are possible vectors of malicious attacks\cite{b2}. In nature, these devices are interconnected to collect and exchange sensitive data of owners. In fact, it is predicted that by 2025 the amount of data produced by IoT devices will be 79.4 zettabytes \cite{b3}.

\noindent 

As the number of the interconnect devices grows on the network, malware attacks targeting them increases \cite{b2,b3,b4,b5,b6}.
In recent years, various modern technologies driven by machine learning have been used for solving complex problems due the rapid development of hardware and heterogeneous devices (including edge and IIoT devices) as well as the widespread deployment of the 6G network that made edge devices gain greater communication and computation capacity for a wider range of applications \cite{b7}. In this regard, machine learning techniques has gained success in developing intelligent systems for intrusion detection in IoT systems. However, they require a large amount of high-quality data for model training in real-world scenarios for their success \cite{b8}. To satisfy this requirement, huge amounts of data is required to be sent to the central entity for model training and modeling purpose. Moreover, over recent years, because the implementation of machine learning as well as deep learning-based techniques gained success for developing intelligent systems, they have been adopted for the IoT/ IIoT security field for cyber-attacks.

 Those techniques use the data gathered by edge devices to detect behavior changes in the network packets. Once behavior sources are selected and monitored, the next step to achieve a successful malware detection is to process the data and generate device behavior fingerprints. Currently, for malware detection solutions, most of existing ML/DL approaches depend on a central server, which gathers data from various IoT devices, and then trains global models. After this step, produced models are sent to each client that is concerned. In other situations, these devices transmit their live test data to the server for behavior evaluation and malware detection \cite{b9}. However, for situations where IoT device’s behavior include confidential data that would impact the security and privacy of owners when captured by attackers, the data owners might be reluctant to use  ML/DL techniques might due to data transmission over the network. Luckily, to ensure data privacy and securing IoT networks, the decentralized ML/DL based anomaly detection models named federated learning have been designed in \cite{b10, b11, b12, b13}. It facilitates a collective learning of complex models between scattered devices’ data guaranteeing data is not shared with the external entities over the internet. Moreover, FL allows training machine learning models with decentralized data while preserving its privacy by design \cite{b9}.
 
Among the researchers that have worked towards addressing the issue of securing IoT network is the authors in \cite{b9} where, by using a FL autoencoder model combined with multi-layer perceptron to detect malware in IoT devices, they complied with data privacy. They compared the result obtained from detecting malware in IoT devices with a centralized approach. Moreover, they have worked on fighting against adversarial attacks when the dataset is processed at the edge because adversarial attacks may affect federated training in terms of both communication rounds and accuracy \cite{b14}.
 
Nguyen et al. \cite{b10} also propose an Intrusion Detection System based on anomaly detection for IoT. Different security gateways, each monitoring the traffic of one a particular device type, locally train Gated Recurrent Unit model and send it to an IoT security service for aggregation. 
Though those FL approaches and others have started to be utilized for anomaly detection, they are addressing the problem of detecting and classifying anomalies in a edge networks on datasets that do not represent the modern IoT/IIoT edge devices. Utilizing new real-world datasets that represent the modern attacks which may resulted from the evolution of IIoT architecture that leads to different novel attacks for training ML/DL even FL model is as important as designing them to be able detected anomalies with higher accuracy and precision.As mentioned above, to comply with data security, the adoption of privacy preserving techniques such as federated learning (FL) for securing billions of vulnerable IoT devices, has started to gain popularity. FL preserves the security for sensitive data held by IoT devices by training a model without transmitting sensitive data over the network. Though, the researchers have been designing and developing decentralized machine learning for anomaly detection in industrial internet of things networks, training these models on the right dataset is a crucial part. Most of those research has been tested on datasets that are not representing modern IoT/IIoT devices, and these devices’ architecture have been growing as the technology advances.

As a result of emerging technologies, new attacks also evolve making it hard to detect attacks that are specific to modern IIoT edge devices with high accuracy. Thus, it is very important to use datasets that closely mirror real-world IoT/IIoT applications. It also is worthy to note that when applying federated learning on IIoT networks, datasets are generated across heterogeneous edge devices and might be unlabeled. To overcome this issue, unsupervised decentralized machine learning is the solution for training models to detect those cyber-threats using unlabeled data.  The authors in \cite{b15} have applied the unsupervised federated learning for detecting anomalies in IoT devices without compromising the data privacy, however, they used dataset that does not contain IoT and IIoT traffic. Moreover, other researchers such as Rey et al. \cite{b9} applied the unsupervised privacy preserving model training on dataset that contains IoT network flows, but the dataset includes only cyber-attacks from two botnets, and there is no IIoT traffic involvement. Therefore, it is difficult to identify and detect other types of malicious IIoT attacks relevant to IIoT security applications, such as man in the middle. Though the work of the authors in \cite{b16} has leveraged the benefits of FL on modern IIoT traffic datasets and have achieved a better accuracy, nevertheless, they didn't consider the unsupervised need of anomaly detection in distributed datasets. Additionally, as most other works related to applying FL for IoT/IIoT security, they used vanilla federated learning, which by default is synchronous and assumes that devices have the same computing power. 

\subsection{Motivation  }
\label{sec:into_Motivation}
Besides data owners being reluctant to avail their data,  in recent years, data legislators and regulatory agencies have been putting a lot of effort into making sure that personal data is secured, making data gathering even more difficult \cite{b3, b17, b18}.  Industries also are reluctant to share their data due to peer competitions, privacy concerns, and other potential concerns \cite{b6}. Moreover, data legislators and regulatory agencies such as General Data Protection Regulation (GDPR) have been putting a lot of effort into making sure that personal data is secured. As a result, gathering data from different reliable sources, especially for IoT devices that hold critical data for their owners, is almost impossible if not costly.

For this reason, the ML models built by utilizing distributed datasets are the possible solutions. One of this approach is possible by applying Federated learning (FL), which is a novel paradigm that enables multiple parties to collaborate on ML model training without requiring direct data access  \cite{b6}.
Furthermore, it is believed that as the edge device architecture evolves, the attacks against modern IoT/IIoT gadgets evolve making the network cyber-attack flow behaviors change leading to high false negatives if the intelligent intrusion detection systems used have been trained and tested on datasets that do not mirror modern attacks.  Thus, evaluating FL model have to be performed on modern datasets that reflect modern IIoT attacks in an unsupervised manner as well because labeling traffic for every devices is not possible. 

\subsection{The research Objectives }
\label{sec:into_esearch Objective}
Privacy preserving distributed ML models that cope with data heterogeneity for anomaly detection and classification in modern IoT/IIoT networks while considering that edge devices are computational heterogeneous and that the amount of data held on each device is variable still need more effort. Considering that, the aim of this research is to use the state-of-the-art distributed ML, which is federated learning by designing and evaluating a fair communication-efficient privacy preserving  distributed (FL) deep auto-encoder model that addresses issues caused by heterogeneous data distribution. The following are breakdowns of the objective steps:
\begin{itemize}

\item	Explore the types of attacks targeting IoT/IIoT devices layers specifically at the network layer.

\item 	Investigate different approaches used for securing IoT devices at the network layer such as machine learning as well as deep learning approaches and their shortcomings as IoT/IIoT devices’ security is concerned.

\item Investigate how Federated Learning (FL) has been adopted for intrusion detection in IoT/IIoT devices and explore the possibility of using its privacy enhanced feature to detect malware with improved training convergence and model accuracy that affecting the modern IoT/IIoT devices due to the emerging technologies.

\item Train and evaluate on a recent dataset that have recent IoT/IIoT attacks by using  unsupervised  federated learning  setting for malicious network flows of IoT/IIoT devices.

\end{itemize}

Most prior federated learning approaches for anomaly detection in IIoT networks use and evaluate the designed model on  datasets that do not represent the modern internet of things. In addition to that, each device in real life is assumed to be producing its own dataset that is not labeled, and those devices might have different computing power as well as use different types of networks for sending the updates weights to the server. Given that most prior research have applied FL for intrusion detection in IoT networks, our project aim was planned as follows:

\begin{itemize}

\item	first, trained and evaluated an unsupervised  deep auto-encoder ML learning on a modern IoT/IIoT dataset.

\item 	designed, trained and evaluated an unsupervised deep auto-encoder federated learning FedAvg on a modern IoT/IIoT dataset.

\item Finally, assessed federated learning FedAvg performance in regard to unsupervised centralized approach both being trained using the recent real-world dataset.

\end{itemize}

\section{Literature Review}
\label{ch:lit_rev} 
To tackle the security issues targeting the IoT devices network, Intrusion Detection Systems (IDS) has been the preferred approach. Intrusion detection methods build a model by studying the behavior of the normal samples through their features, and any deviation can be detected as suspicious action on the device \cite{b19}.  IDSs are normally classed into signature-based or anomaly-based approaches. Signature-based IDSs identify intrusions and protect the system by utilizing previously learned rules from known attacks. Anomaly-based IDSs monitor network traffic and compare the traffic with previously learned patterns to spot malicious activities. In comparison, anomaly detection methods have shown to be more effective to recognize known and new attacks but need frequent updates to detect new attacks \cite{b12}. Over recent years, the implementation of machine learning as well as deep learning-based techniques have gained success for developing intelligent systems, and they have been adopted for the IoT security field for cyber-attacks. 

\subsection{Centralized machine learning approach}
\label{ch:Centralized_machine_learning_approach}

After the well known world DDoS attack caused by  IoT malware Mirai , which was used to propagate the biggest DDoS  attack in 2016, there has been a proliferation of IoT malware attacks following the leakage of its source code. Since then, various research has been done to address the issues. Among those solutions are ML-based and DL-based, which are widely utilized in cyber security \cite{b20}. Regarding the computational issue of IoT devices, attempts to build ML light-weighted algorithms that can be used to secure IoT by detecting malicious traffic and nodes have been done. Kumar et al. \cite{b21} proposed a module for detecting the network activity of IoT malware in large-scale networks called EMDA during the scanning /infecting phase rather than during attack. It consists of machine learning (ML) algorithms running at the user access gateway which detect malware activity based on their scanning traffic patterns, a database that stores the malware scanning traffic patterns and can be used to retrieve or update those patterns, and a policy module which decides the further course of action after gateway traffic has been classified as malicious.  To reduce the memory space required, the classification is performed on IoT access gateway level traffic rather than device-level traffic as working on aggregate traffic is faster. two classes of gateway-level traffic were classified  as benign and malicious using Naive Bayes, k-NN (k-Nearest Neighbor) and Random Forest algorithms. Though the accuracy was above 88\%, it still needs to be improved.  In addition to that, there is an added issue of communication latency, and they did not consider the effect of false alarm issues. 

Fernandes et al. \cite{b22} tackled the issues by proposing an integrated detection mechanism against DDoS attacks for the IoT network environment, called Smart Detection-IoT (SD-IoT) system. A solution that uses machine learning to classify IoT network traffic and detect denial of service attacks by only analyzing the IP/TCP header of network traffic samples, thus not com- promising data privacy. The detection module for an IoT controller uses Machine Learning (ML) techniques to classify network traffic. The system was designed in the Software-Defined Networks (SDN) context and evaluated on an emulated platform using three actual and well-known datasets.  Variable selection was performed using the Recursive Feature Elimination and Cross-Validated techniques. This method used three well-established machine Random Forest (RF), Logistic Regression and Extreme Gradient Boost (XGB) algorithms. The accuracy is higher, however the False Alarm Rate metrics scored almost 6\%, which is a problem that needed to be addressed. 

Hafeez et al. \cite{b5} to address the issue, they presented IOT-KEEPER, a lightweight system which secures the communication of IoT. It is an edge system capable of performing online traffic classification at network gateways by utilizing unlabeled traffic data for model training. It is also capable of detecting malicious network activity in real-time. IOT-KEEPER uses. It uses a combination of fuzzy C-means clustering and fuzzy interpolation scheme to analyze network traffic and detect malicious network activity. During the experiment, for malicious traffic, they used Mirai infected IoT devices and Raspberry Pis. Once malicious activity is detected, IOT-KEEPER automatically enforces network access restrictions against IoT devices generating this activity and prevents it from attacking other devices or services. Their technique achieved high accuracy ($\approx$ 0.98), low false positive rate ($\approx$  0.02) for detecting malicious network activity and raised a few false alarms (0.01). Even though they assert that they solved the problem of computation and false alarms, it still consumes additional memory for storing device data and security policies. Moreover, since IOT-KEEPER only analyzes new flows in the network, the resulting increase in CPU utilization is small, however, in the worst case, 0\% cache hit rate, average CPU load increases up to 83.3\%. 
\subsection{Federated learning approach}

Though those researchers have improved the security of IoT devices, achieved higher accuracy, minimized the latency as well as computation cost by bringing the load to the edge, they are still issue of the shared unsecured sensitive data to the edge over the internet. For that reason, a decentralized ML method termed Federated Learning (FL) is introduced in \cite{b13}. This approach facilitates a collective learning of complex models between scattered devices guaranteeing data is not shared over the internet. As define by Nguyen et al. \cite{b10}, each device utilizes its own local data for training and collaborate with others. Only the local computed updates, such as the model’s weights, are sent to the FL server for aggregation. For achieving a high accuracy, the steps are repeated in multiple rounds. A federation of the learned and shared models performed on a central server to aggregate and share the built knowledge among participants helps to improve the entire system \cite{b6}.  Thus, FL is seen as an important ML approach to secure IoT devices. For this reason, Nguyen et al.\cite{b10} propose an FL Intrusion Detection System based on anomaly detection for IoT called DIOT. It performs dynamic detection of any unknown attacks that deviate from the legitimate behavior of the device, since it only models legitimate network traffic. Different security gateways, each monitoring the traffic of one particular device type, locally train the Gated Recurrent Unit model and send it to an IoT security service for aggregation. Such a system is able to detect novel attacks. However, its accuracy for malware detection for devices in standby mode is still low (88.96\%). Moreover, having a central FL server for aggregating the model through improved security of data, the participating devices’ communication limitations as well as data heterogeneity are some of the big challenges for model convergence. Few works have been done trying to solve the problem. Rahman et al. also proposed a Federated Learning based scheme for IoT intrusion detection that maintains data privacy by performing local training and inference of detection models. They evaluated their model on NSL-KDD dataset and then compared the results with the centralized approach. With data legislators and regulatory agencies imposing data privacy designing systems using decentralized approach is the way to

\section{Methodology}
\label{ch:method} 
For implementing the proposed methodology, some steps must be considered.  Initially, we defined the way synchronous federated learning differs from fair synchronous federated learning and clearly outline the type of the edge devices that need to be involved and secured. Then, the possible  way a device will interact with the server without uploading its dataset is also designed. The architecture and type of unsupervised algorithm to be used for anomaly detection is also  designed. That is a deep auto-encoder for a federated environment. Finally, the framework evaluation  is performed on Edge-IIoTset\_2022 by considering the presence of Independent and Identically Distributed data across clients. This helps in describing the way the application is designed and evaluated for observing the unusual activities during the run-time.

\subsection{Synchronized Federated Learning Optimization}
Normally, for federated learning schemes, a global deep learning model resides on the server-side, whose weights are initialized randomly. Then, the global model is sent across the edge (IoT/IIoT) devices. Upon receiving the initial model, all the participating edge devices parallelly train the global model with the dataset that they held locally on their storages. The updated weights from each participant are sent back to the server for aggregation. The server waits for some amount of  time for all active participants to send their weights before averaging.  Once the global model averaging is done, the averaged weights are then again sent back to the edge devices for the next round of training \cite{b23}. This is repeated in several rounds until the global model achieves the desired results. The training continues until we have reached the desired accuracy.


\begin{center}
    \includegraphics[width=.8\columnwidth]{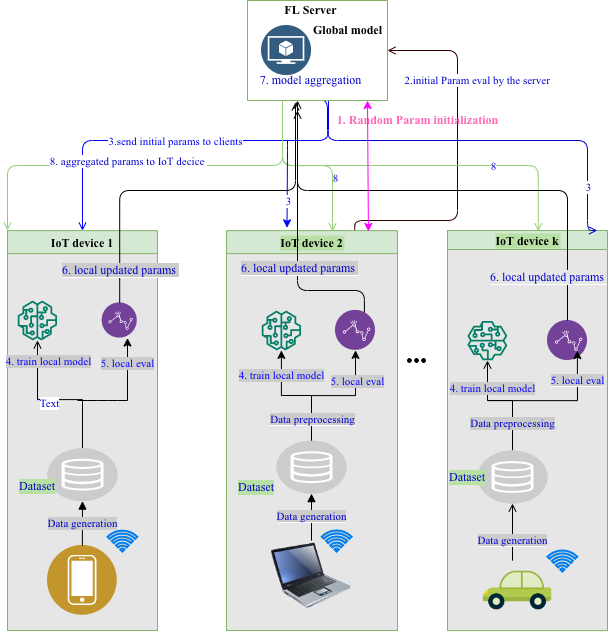}
\end{center}
\begin{figure}[h]
\caption{ \emph{Illustration of federated learning from the step one until step 8 where aggregated weights are sent back to the local devices for continuation of the process until the desired accuracy is achieved. Note that from step one to step 8 is one round. The training is performed for many rounds depending on the problem at hand.}}
\centering
\end{figure}

As shown in Algorithm Federated Averaging algorithm 1 FedAvg, at each global iteration,a subset of the devices are selected to run gradient descent optimization (e.g., SGD) locally to  optimize the local objective function \( F_k\) on device k. Then, these local devices communicate their local model updates to the server for aggregation. With heterogeneous local objectives  \( F_k\), carefully tuning of local epochs is crucial for FedAvg to converge. However, a larger number of local epochs may lead each device towards the optima of its local objective as opposed to the central objective. Besides, data continue to be generated on local devices which increases local gradient variations relative to the central model. Therefore, for the future,we plan to to incorporate a fairness approach to restrict the amount of local deviation by penalizing large changes from the current model at the server. We explain this in detail in Section \ref{sc:Fairness and communication-efficiency}.

Moreover, most synchronized federated optimization methods have a similar update structure as FedAvg. One apparent disadvantage of this structure is that, at each global iteration, when one or more clients are suffering from high network delays or clients which have more data and need longer training time, all the other clients must wait or this devices is ignored in training the model together leading to the model being biased to those who are faster and able to send their models on time.  This is where a communication efficiency and fairness FL optimization approach comes in, which is explained in section \ref{sc:Fairness and communication-efficiency}.


\begin{figure}[h]
  \includegraphics[width=.45\columnwidth]{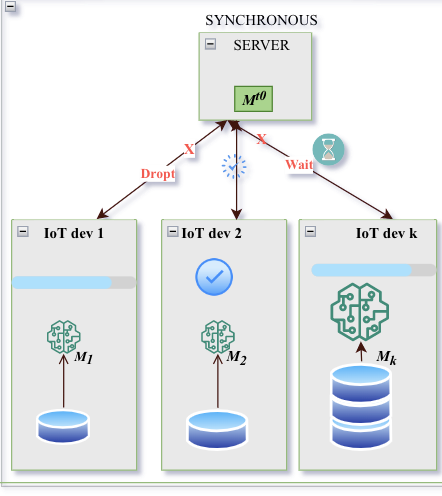}\hfill
  \includegraphics[width=.45\columnwidth]{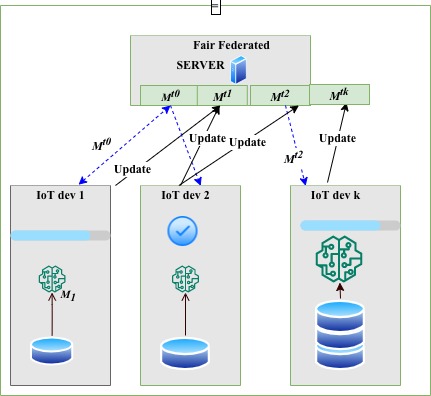}

  \caption{Illustration of unfairness in FL model update.}
\end{figure}

In synchronous IoT Device 1 is a straggler either because of the internet or low battery power. IoT Device K needs extra time to train the model. For the Synchronous FL the central server must wait or drop it to perform the aggregation.

\begin{algorithm}
	\caption{Federated Averaging Algorithm}
	\begin{algorithmic} 
	\Require {$K$ number of devices: $k=1,2,\ldots K$,  and learning rate $\eta$,\
	   $S_t$ :subset of selected clients, $T$: communication.}
	  \textbf{Initialize:}
        \State $W^{0}$$\leftarrow$\textit{random initialization} \\
	    \State \textbf{\underline{Procedure at Central Server}}
		\For {$ t= 0,\ldots, T -1$} 
		\State Server samples a subset $S_t$ of $K$ of available devices.
		\State Server sends $W^t$ to all selected devices.
		\State Each  device $k$ updates $W^t$ for E epochs with step-size $\eta$ to obtain  $W^{t}$ 
		\State Each activate selected device sends $W^{t}$ back to the server
		\State Upon receiving $W^{t}$ of active clients, server computes the average using  Eq.: 3.1.

		\State Perform step to update the model weights.
		\State Server sends $w^{t+1}$ to all selected devices
        \EndFor
	\State \textbf{ \underline{Procedure of Local Client $k$ at round $t$}}
	\State Compute the $w^{t}$ at every round.
	\State Each selected device \(k \) sends $w^{t}$ back to the server
\end{algorithmic} 
\end{algorithm} %



\subsection{The approach for how  devices interact with the server}
\label{ch:The approach for how  devices interact with the server}

The server starts to update the global model \(M\) after receiving updates from one device, and  all clients maintain their own model copies \(M^K\) of \(M\) in the memory. The server starts aggregation after receiving at least two clients' update and performs feature learning on the aggregated parameters to extract a cross-client feature representation \cite{b24}. Then the server starts the next iteration and distributes the new updated aggregated parameters to the ready clients. 

\subsection{Possible approach for fairness.}
\label{ch:Proposed approach }

For the achieving fairness in FL methods, the server and edge devices communicate by sharing weights. The server sends aggregated weights value to the devices, then the devices perform step, do the local training, return the updated weights send them to the server for server aggregation. For the server optimization purposes such as avoiding being idle, the server computes the averaging as longs as there are two or more active clients. The description of the approach consist of two procedures. One at server side and another at local edge device.
The procedure at the central server, which acts as the coordinator is describe in algorithm 2. The server randomly selects one available client. The selected device then sends an initial computed weights \(w^0\). Upon receiving the initial weights, the server samples a subset of \(S_t\) clients to send the weights to. Then, the server computes \(\nabla_k(w^0)\) and store it in a gradient history \(GH\) list. This list of updated weighted gradient descent values is used for consistency when deciding to perform averaging and calculating the relevance scores of the gradients as explained in the equation 5 and 6. Since the devices might send their updates in different iterations, some might have been involved in the current iteration or it is first step, we need to keep the model consistent.  The value of \(w^0\) is from one randomly selected client at first iteration. For the next step, it receives one by one the \(w_k^{t+1}\) of selected participants and starts computing averaging as the clients send them depending on the step of the \(i^{th}\) iteration \(T\).  If the number of participating clients is greater than two and not less than the subset of selected clients \(S_t\), it means the clients have been sending their updated \(w_k^{t+1}\). Then, perform aggregation using equation 1. But, this equation will be incorporated in equation 3 for fairness purposes that are explained in \ref{ch:communication-efficiency_Federated_Learning_design} section. 

\begin{align}
    w^t= \frac{1}{S_t}\sum_{k=1}^S w^t
\end{align}  

\subsection{Fairness and communication-efficiency:}
\label{sc:Fairness and communication-efficiency}

Given  that in federated networks data are often non homogeneous in respect of both size and distribution, and model accuracy may generally vary, an efficient  optimization approach for federated learning method that encourages more uniform distribution of the model accuracy  across the participating devices is very important. To avoid for example, the aggregated model to be biased towards devices that have larger numbers of data points, or faster devices that keeps sending data point to the server, Li et al. \cite{b25} proposed q-FFL that minimizes an aggregate re-weighted loss parameterized by \(q\) such that the devices with higher loss are given higher relative weight. In their research, each selected device \(k\) updates \(w^t\) for E epochs of SGD on \(F_k\) with step-size $\eta$ to obtain  \(w^{t+1}\). For a strong form of uniform continuity each local device k computes  the upper-bound of the local Lipschitz constant as follows $LF_k(w)^q + qFk(w)^{q-1} \| \nabla F_k(w)\|^2 $.
In each iteration, the weights (step-sizes) $\nabla w_t^k $ are inferred from the upper bound of the local Lipschitz constants. They also propose a heuristic $ h_t^k $ where they replace the gradient of Federated SGD  steps with the local updates that are obtained by running SGD locally on device k.

\noindent

Instead  of so much computation being done at the client device side, we will shift the work at the server side, which have much resource for accelerating the process, and only shared over the network the weight value of a given active clients.  As a result,this will intuitively reduce the communication as well as computation cost because the added time of computing necessary parameter will be done at the server side and the parameter shared over the network is only weight values. 

\subsection {How can fairness be achieved ?}
\label{ch:How can fairness be achieved ?}

To Achieve fairness among participants' model accuracy, the ideal way would be to \textbf{\textit{reweight}} the objective function.  This means, for devices with poor accuracy, higher weights is assigned to them. In this way, the distribution of accuracies in the network will become more uniformity as we train the model. This is the approach of (q-FFL) for achieving fairness. The re-weighting must be done dynamically, as the performance of the devices depends on the model being trained, which cannot be evaluated a priori \cite{b26}.

 Given local non-negative cost functions \(F_k \) and parameter \(q > 0\),  the q-Fair Federated Learning (q-FFL)  objective is as follows:

\begin{equation} \label{fairness}
    minf_q(w) = \sum_{k=1}^m \frac{pk}{q + 1} F_k^{q+1}(w)
\end{equation} 

\noindent
\(F_k^{q+1}\) denotes \(F_k(.)\) to the power of \((q+1)\). The parameter \(q \) tunes the amount of fairness that is desired for better performance. If \(q \) = 0, it means no fairness beyond the classical federated learning. A larger \(q \) means that the focus is on devices with higher local losses, \(F_k(w)\), hence training accuracy distribution uniformity is imposed and potentially reaching fairness according to the definition of fairness given by Mohri et al. in \cite{b26}. According to Li et al. in \cite{b25} a \(f_q(w)\) a large enough \(q\) reduces to classical minimax fairness , as the device with the worst performance (largest loss) will dominate the objective \cite{b26}.

\subsubsection{\textbf{Improving communication-efficiency:q-FedAvg}} 
\label{Improving communication-efficiency:q-FedAvg}
Communication-efficient strategies that use local stochastic methods such as FedAvg remarkably improve convergence speed of the model. However, in the scenario where \(q > 0\), the \(F^{q+1}_k\) term is not an empirical average of the loss over all local samples due to the \(q + 1\) exponent. As a result, local SGD as in FedAvg is prevented to be used . To address this, Li et al. \cite{b25}  proposed to generalize FedAvg for \(q > 0\) using a more sophisticated dynamic weighted averaging scheme. The weights (step-sizes) are inferred from the upper bound of the local Lipschitz constants of the gradients of \(F^{q+1}_k\) , similar to q-FedSGD. To extend the local updating technique of FedAvg to the q-FFL objective (\ref{fairness}), they proposed  heuristic steps defined in equation (\ref{heuristic steps}). Similarly, q-FedAvg is reduced to FedAvg when \(q \)= 0.

\subsection{Central server node activities for Fair FL}
\label{ch:Central server node activities for Fair FL}

This part highlight how learning at the server is done. The server receives gradients as they come in. From them, the model weight \(w\) is deducted. Then, after every global iteration the central aggregated model \(w^{t+1}\) is computed by the server. At the first iteration where the server get a random weight value \(w_k\), the initial model weight \(w^t\) is then derived. At global iteration \(t+1\), suppose the central server gets updated gradient from client \(k\). Let \(w^{t+1}\) be the server model, \(w^{tk}\)  be the local model of client \(k\) at iteration \(t\), \(g_k\) be the gradient on the local data of client \(k\), \begin{math} \eta_k^t \end{math}  be the learning rate of client \(k\) and \(D_K\) = \(d_1\) + ... + \(d_k\)+ ... + \(d_K\) be the present total number of data points of participating devices where \(D_k\) denotes the local dataset of the \(k^{th}\) client, and \(n_k\) is the size of \(D_k\). When computing the aggregate of the updates from participating devices by the server, then, the global weight value is updated

\noindent We first compute the model weights $\nabla w_t^k $, $\nabla_t^k $ and $ h_t^k $  before computing $ w^{t+1 }$. As mentioned above, most of the  computing activities are done at the sever side. After computing the gradient descent value, the sever will also compute the following formulas adopted from \cite{b25} for fairness purpose:
\begin{equation} \label{heuristic steps}
\begin{split}
    \nabla w_t^k &= L(w^t - w^{t+1})\\
    \nabla_t^k  &= F^q_k(w^t)\nabla w^t_k\\
    h_t^k  &= qF^{q_1}_k (w^t) \| \nabla w^t_k\|^2 + LF^q_k(w^t)
\end{split}
\end{equation} The aggregation of weights at the sever, which will sent over the internet to the active participating clients, is as follow: 
\begin{equation}\label{4}
\begin{split}
     w^{t+1 }& =w^t - \frac{\sum_{k\in S_t}\nabla_k^t}{\sum_{k\in S_t}h_t^k}
\end{split}
\end{equation}

\noindent
Because the server computes average of the weights  as they come, their current aggregated value of produced gradient by the server might not reflect the ones from struggling devices. For example  as it is illustrated in Figure 2, the time the central server  receives the value sent  by the straggling devices (e.g., Client 2), it has so far updated the global model twice. This is one of the sources of  inconsistency in the asynchronous update scheme when it comes to obtaining model updates from the server. This model inconsistency cannot be avoided in the real world settings  because of data and system heterogeneity, or network delay. We considered this issue on the server side and used the strategy of feature scoring using in natural language processing to solve it.


\subsection{Feature Representation Learning on Server.}
\label{ch:Feature Representation Learning on Server.}

As mentioned above, the model weights are aggregated as they come except when we have less than two clients, the current value of produced global weights by the server might not reflect the ones from struggling devices. Thus, on the server side, we extract a cross device feature representation. This is done by using technique in natural language prepossessing domain where prediction of the relevance score of each context vector is used \cite{b27}. This relevance score measures how relevant the \(i^{th} \) context vector of the source sentence is in deciding the next symbol in the translation. Adopting if for our problem, we want to apply it to be able to decide the weigh value of the next \(i^{th} \) iteration using the current available weights sent by currently participating clients. We first compute the relevance scores \textbf{\begin{math} \alpha{t^i}\end{math}} which is normalized before being used. 


\begin{equation} \label{5}
\begin{split}
     \alpha^{i+1} \leftarrow  \frac{\vert\emph{exp}^{(w^{t+1})}\vert }{\sum_{j=1}^{T_k }\vert\emph{exp}^{(w^{t+1})}\vert} 
\end{split}
\end{equation}
Then multiply \begin{math} \alpha t^i\end{math} with the averaged weight as it is shown in equation 

\begin{equation} \label{6}
     w^{t+1}= \alpha t^{i+1} * w^{t+1}
\end{equation}
It is important to note that feature representation is done regardless of the number of participating clients at \(i^{th} \) iteration. Even in a scenario where at \(i^{th} \) iteration we have only one client, except the first iteration, we still must perform feature representation for model consistency.
After computing that, we will have to update the trace $GH=[\nabla_1^t, \nabla_2^t,……, \nabla_k^{ST}]$, perform step to update the server model parameter.Finally, send the current gradient descent to clients.
Procedure at client side is to compute the weight  and perform training until the last iteration when the desired accuracy is achieved. 

\begin{algorithm}
	\caption{FairFedAvg}\label{alg1}  
	\begin{algorithmic}
	   \Require {$K$ number of devices: $k=1,2,\ldots K$,  and learning rate $\eta$,\
	   $S_t$ :subset of selected clients, $T$: communication,relevance scores $\alpha_t^i$ rounds.}
    
	\State \textbf{Initialize:}$W^0$ and gradient history list  $GH$ = [], $\nabla\_k^t$ = [], $hs\_ffl $= []
	    \State \textbf{\underline {Procedure at Central Server}}
		\For {$ t= 0,\ldots, T -1$} 
		\State Server randomly samples a subset $S_t$ of $K$ of available devices.
		\State Server sends $W^t$ to all selected devices.
		\State Each selected device $k$ updates $W^t$ for E epochs with step-size $\eta$ to obtain  $W^{t}$ 
		\State Each activate selected device sends $W^{t}$ back to the server
		
		\State Upon receiving $W^{t}$ of active clients, server computes the  Eq.: \ref{5} to get $\nabla_k^t$ and $hs\_ffl $.
		\State Stores gradient value in $GH$ = [$\nabla_k^{t}$]
		\State Server updates  $w^{t+1}$ as follows:
		
		\If {$len(\nabla_k^t )>2 \ and \ S_{t+1} =S_{t-1}$} 
	    \State 	\begin{equation*}
        \begin{split}
         w^{t+1 }& =w^t - \frac{\sum_{k\in S_t}\nabla_k^t}{\sum_{k\in S_t}h_t^k}
        \end{split}
        \end{equation*}
        \Comment{get the update value of $w^{t}$ from selected devices.}
    
		\EndIf
		
		\If  {$len(\nabla_k^t )>2\ and \ S_{t+1} < S_{t-1}$} \
		\State Update  the value of $w^{t+1}$ with feature learning 
	    \State $w^{t+1}= \alpha^{t^{i+1}} * w^{t+1}$
	     \EndIf
	    \If {$ S_{t+1} < S_{t}\ $}\
		\State Update  the value of $w^{t+1}$ with feature learning 
	    \State $w^{t+1}= \alpha^{t^{i+1}}  * w^{t+1}$
	    
		\Else \ 
		\State $w^{t+1}= w^{t}$
	    \EndIf
		\EndFor
		\State Update the trace $GH=[\nabla_1^t,\nabla_2^t,\ldots,\nabla_k^{ST}]$
		\State Perform step to update the model weights.
		\State Server sends $w^{t+1}$ to all selected devices
	\State \textbf{\underline{Procedure of Local Client $k$ at round $t$}}
	\State Compute the $w^{t}$ at every round.
	\State Each selected device \(k\) sends $w^{t}$ back to the server.
	\end{algorithmic} 
\end{algorithm}

\section{Implemented approach and Response Features}
\label{ch:Iplemented approach and Response Features}
\subsection{Federated Averaging}\label{Federated Averaging}
 Federated learning FL as a distributed machine learning approach performs training of a machine learning model on the device that gathered the data itself. In this scenario, the data is not share over the network for training purpose.  Fedavg as one of FL algorithms permits a model to be copied to participating devices during a training session. The devices could be chosen at random, and a device can be aborted.  The resulting models are sent to the coordinating server and then average models from the devices that finish training. By doing this, Federated learning solves the privacy problem for IoT/ IIoT devices that held sensitive data for the owners. In this project, we implemented this strategy on a recent dataset that represent the modern IoT/ IIoT device networks
\subsection{Communication-efficiency Federated Learning design}
\label{ch:communication-efficiency_Federated_Learning_design}
For our work, the model is initiated on each device, then the server gets the initial parameter from one client at random, and the process follows as usual. To alleviate the burden of to much computation being done on the client side and sharing many parameters on the network, which in turn become congested, most of computations are designed to be done by the server side. The consideration of IoT/IIoT devices in this experiment that can perform some computation such as smart fridge or smart cars is worthy to be clarified.

\noindent Federated Learning has the potential to break the barrier of centralized ML/DL adoption at the edge through better data privacy. However, the heterogeneity among the clients’ compute capabilities, the amount and quality of data can affect the training performance in terms of overall accuracy and convergence time.  We designed a framework that allows fair participation of edge devices based on feature representation. We also considered the presence of Independent and Identically Distributed data across clients.

\subsubsection{ Algorithms Used }

	The deep learning algorithms used in this paper are as:
	
\begin{itemize}
    \item Unsupervised deep auto-encoder algorithm
    \item Synchronous unsupervised federated averaging learning (FedAvg)-deep auto-encoder
\end{itemize}

\subsubsection{Response Features}
The problem to solve in this project is a binary classification problem, which  means :
\begin{itemize}
    \item Attack or malign network flows
    \item Normal or benign network flows
    
\end{itemize}

\subsection{Deep Auto-encoder Design}
\label{ch:FDeep Auto-encoder Design}
For processing the unlabeled data, an unsupervised deep learning-based approach known as auto-encoder is used. The auto-encoder is responsible for turning the unlabeled traffic data into labeled traffic data by extracting features from the unlabeled data. The proposed auto-encoder architecture is shown in figure 3. Auto-encoder takes place in two phases, i.e., encoding and decoding\cite{b28}. The encoding step maps the input X feature set of a dataset and then transforms it to hidden representation as in equation \ref{5}. As the equation \ref{5} illustrates, the hidden representation is denoted by y while W  and B  are the weight and bias associated with the encoder. For the step of a decoder, the decoder takes the y as input and reconstruct the original data from hidden representation as it is demonstrated in equation 3.6, where $W$ and $B$ are the weights and bias associated with the decoder and $z$ is the output.

\begin{align}
    y = f(x)_{encode} = \Phi(W_\phi * x + B_\phi )  
\end{align}

\begin{align}
    z = f(y)_{decode}= \theta(W_\theta  * y + B_\theta)    
\end{align}

\noindent The deep autoencoder used for training and evaluation is composed by the input layers of 66
neurons, 3 hidden layers of 128, 64 and 32 neurons each followed by a dropout probability p
set to 0.2, and the encoded layer(bottleneck) of 16 neurons.

\begin{center}
    \includegraphics[scale=0.6]{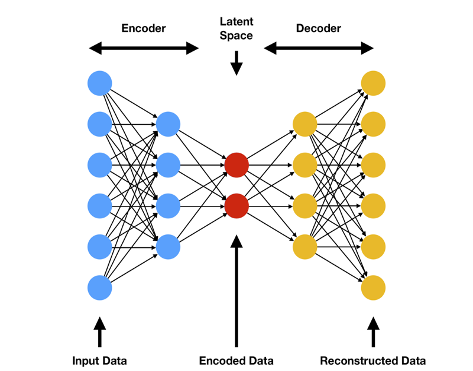}
\end{center}
\begin{figure}[h]
\caption{ \emph{Auto encoder design}}
\centering
\end{figure}

\subsection{Data Exploration}
\label{ch:Data Exploration}

\subsubsection{Dataset used.}
\begin{itemize}
    \item Dataset EdgeIIoTset description   
    
\end{itemize}
Edge\_IIoTset \cite{b28} is a new cybersecurity dataset for IoT and IIoT applications. It is designed to test the performance of intrusion detection systems. It was chosen based on the reason of being produced after involving several layers containing new emerging technologies. Those technologies fulfill the key requirements of IoT and IIoT applications, such as, ThingsBoard IoT platform, OPNFV platform, Hyperledger Sawtooth, Digital twin, ONOS SDN controller, Mosquitto MQTT brokers, Modbus TCP/IP, …etc. moreover, it also is realistic for testing machine learning in federated learning manner.  The data is generated from various IoT devices such as temperature and humidity sensors, Ultrasonic sensor, Water level detection sensor, pH Sensor Meter, Soil Moisture sensor, Heart Rate Sensor, Flame Sensor, etc. The dataset includes 14 types of attacks belonging to the following categories:
\begin{enumerate}
    \item 	DoS/DDoS attacks,
    \item Information gathering,
    \item Man in the middle attacks, 
    \item Injection attacks, 
    \item and Malware attacks. 
\end{enumerate}

The dataset is divided into two parts for machine learning and deep learning models. In this work we chose to use the deep learning dataset. The table 1 shows the statistics of the dataset
\begin{table}[H]
    \centering
    \caption{Edge IIoTset dataset statistics}
    \label{tab:_ex_tab}
    \begin{tabular}{llr}     
        \hline
        \multicolumn{2}{c}{Number  records} \\
        Type    &  Category & Quantity  \\
        \hline
        Number of normal records    & -   & 300000   \\
        Number of attacks      & ransomware     & 2000   \\
                 & password    & 2000   \\
                 & scanning     & 2000   \\
             & injection   & 2000   \\
              & xss   & 	2000   \\
               & dos    & 2000   \\
                & backdoor   & 2000   \\
                 & ddos   & 2000   \\
                 & 	mitm   & 1043   \\
        \hline
    \end{tabular}
\end{table}
 
\subsection{Data Preprocessing}
\label{ch:Data Preprocessing}
An autoencoder is trained by minimizing the Mean Squared Error (MSE) between the reconstructed features and the input \cite{b23}. For this reason, to detect anomaly, an autoencoder is supposed to be trained on a dataset containing normal network flows so that it can be able to learn how to reconstruct normal flows as outputs with a low reconstruction error when it is faced with benign data and a high reconstruction error on other abnormal data. That’s why for the centralized model, we first separated the normal flow from anomalous network flows. Then, from the normal data, we make a split of 80\% for the train set and 20\% for the validation set. The remaining dataset containing abnormal network flows was used for testing by computing the same 20\% data points as used for validation. The training process is used for computing a threshold, which is set depending on the reconstruction error of training on normal data. While performing testing, if input data reconstructed error is higher than the computed threshold, then it is classified as abnormal otherwise the classification result is decided as a normal flow. 

On the other hand, for FL, we used the same strategy but, the data points were assigned to two clients by simulating the presence of different data distribution on devices (Non-IID character) using a function present in Flower framework, which is called latent Dirchlet Allocation. This function can produce some data distribution for classification problems from one label data to purely uniform distributed data depending on the concentration value called alpha -- as is explained in \cite{b29}. The lower the value of the alpha the more the distribution is different and diverse. On the other hand, if the distribution is high like where alpha is 1000, the dataset is normally distributed. For our experiment we used the alpha value of 10 which is enough to have non\_IID dataset.


\subsection{Evaluation Metrics}
\label{ch:Evaluation}
The following metrics are used in evaluating the performance of the proposed FL anomaly detection.
\begin{enumerate}
    \item True Positive (TP): the true malicious network data flow is detected.
    \item False Positive (FP): normal network data is detected as malicious.
    \item True Negative (TN): benign data is classified as a malign activity. 
    \item False Negative (FN): When malicious activating is not detected as a non-malicious activity. 
    \item Threshold (TR): The level of mean square reconstructed error for discriminating normal data flows from and malicious data. 
    \item Mean Squared Error (MSE) 
    \item Standard Deviation (STD) 
    
\end{enumerate}

\subsection{Training  Metrics}
 The main goal of the autoencoder is to try to minimize the loss function, which is the mean squared error between the original input and the reconstructed output. The following metrics are used for training:
 \begin{itemize}
     \item Threshold is defined as the mean and standard deviation of the training MSE reconstructed loss. For the experiment, for centralized approach, after the training process is finished, a threshold is computed  using the equation (9)  and fixed at a certain number. This is based on computing mean and standard deviation of the reconstruction error of normal network flows. When testing, if a certain data sample mean squared reconstruction error is  above  the fixed threshold, it is classified as positive, otherwise it is considered as a normal network flow. In this experiment we computed the threshold using a formula that was used by Rey et al.[23]. The threshold is calculated for each client \(k\in K\).
     \begin{align}
       {thr_k}  = {mean(MSE(D_k^{Thr}; w_k))} + {std(MSE(D_k^{Thr}; w_k))}  
     \end{align}
     where \(MSE(D ; wk)\) is the mean squared reconstruction error computed with model parameters \(wk\). 
     For distributed approach(Federated learning), each training round we computed the threshold, and during testing we set the least threshold among  all rounds' thresholds. The reason for this idea is to reduce the overall false  positive.
    \item 	False positive rate is defined as the sum of the normal data predicted as greater than the threshold during validation on normal traffic over the total number   of the validation data points. 
    \begin{align*}
        FP_{rate}=\frac{sum\hat{y} > thr_k}{ len(validation)} 
    \end{align*}
    
    \end{itemize}

By utilizing the threshold value, we were able to turn the problem into a binary classification where the output greater than the threshold is positive (1) otherwise negative (0).

\subsection{Evaluation Metrics}
\label{ch:Evaluation Metrics}

The evaluation of the model is based on accuracy, precision, recall, and confusion matrix on the testing set.

 \begin{itemize}
    \item Accuracy is defined as:
     the ratio of number of correct predicted data either true abnormal sample and true benign sample over the total number of all  samples.
    \begin{align*}
        Accuracy=\frac{TP + TN}{ TP + TN + TN + FN  } 
    \end{align*}
    
     \item Precision: the ratio of true detected of abnormal data flow, over the sum of true abnormal data flow and false detection of normal data. 
     \begin{align*}
        PR=\frac{TP }{ TP + FP } 
    \end{align*}
     
     \item Recall is the ratio of true positives over the sum of true positive detection and false detection of normal network data.
     \begin{align*}
        Recall=\frac{TP }{ TP + FN } 
    \end{align*}
     \item F-measure is the weighted average of the precision and recall 
     \begin{align*}
        F_{measure}=2 *\frac{Precision * Recall}{Precision + Recall} 
    \end{align*}

 \end{itemize}

\subsection{Testing Metrics } 
\label{ch:Testing Metrics}
As the evaluation process, the testing part of the model is also based on accuracy, precision, recall. in additional to that, we added a confusion matrix for the testing process for summarizing the performance of our classification algorithm. The results are described  in the \ref{ch:Experiments and Results} subsection.
\section{Experiments and Results.}
\label{ch:Experiments and Results}

\subsection{Framework and Tools used}
\label{ch:Framework and Tools used}
Model training and evaluation will be performed   using Flower framework \cite{b30}, which helps in training a model on data you do not own and cannot see. According to Mathur, Akhil, et al. \cite{b31}, Flower is a novel client-agnostic federated learning framework and one of the key design goals is to enable integrating with an inherently heterogeneous and ever-evolving edge device landscape. The Flower core framework, shown in Figure 1, implements the infrastructure to run heterogeneous workloads at scale. On the server side, there are three major components involved: the FL loop, the RPC server, and a (user customizable) Strategy. Strategy here refers to the federated averaging algorithms (e.g., FedAvg) used for aggregating the model parameters across clients. Clients connect to the RPC server which is responsible for monitoring these connections and for sending and receiving Flower Protocol messages. The FL loop is at the heart of the FL process: it orchestrates the learning process and ensures that progress is made. It does not, however, make decisions about how to proceed, those decisions are delegated to the currently configured Strategy implementation.

\subsection{Experiment configuration}

\label{ch:Experiment configuration}
This section describes and reports the training as well as evaluation results of the proposed approach. To approve FL approach as a solution for securing IoT/IIoT devices network environment with privacy of data, we need to demonstrate its effectiveness. Thus, results will be evaluated and compared against a baseline approach. In this regard, the baseline means centralized machine learning process. Moreover, since we are introducing the non-synchronicity for FL approach, synchronous FedAvg assessment with non-synchronous FedAvg is planned. The experiments are carried out on a laptop with specifications of 128GB of RAM, graphic Intel HD Graphics 6000 1536 MB, operating with macOS 10.15.7 (19H1824). All the data pre-processing was done within Python using public libraries. The deep learning framework used is PyTorch, and the FL was implemented using the Flower open-source framework \cite{b30}.

\subsection{Deep autoencoder Training settings }
\label{ch:Deep autoencoder Training settings}

The deep autoencoder used for training and evaluation is composed by the input layers of 66 neurons, 3 hidden layers of 128, 64 and 32 neurons each followed by a dropout probability p set to 0.2, and the encoded layer(bottleneck) of 16 neurons. We have used the relu activations function in the hidden layer of autoencoder, whereas the Tahn activation function was used in the output layer of the decoder part of the deep autoencoder model.

\subsection{Training Results }
\label{ch:Training Results}
 For the centralized ML, training has been performed for many epochs and it turned out that 50 epochs produce best results compared to other results of more than 50 epochs of training. The batch size of 32 was used, and the settings of Adam optimizer are as follows:
 \begin{itemize}
     \item learning rate of 0.001
     \item scheduler of step size of 1
     \item and gamma of 0.9 
 \end{itemize}

 Figure 4 represents the mean square error training progress as well as the threshold for the normal dataset through 50 epochs

On the other hand, though the settings are the same, However, a larger number of local epochs may lead each device towards the optima of its local objective as opposed to the central objective. Thus, for federated learning, it  was trained and evaluated using 10 epochs each round for each of the two clients for 5 rounds to match the number of the centralized approach. Note that for FL, the threshold for the first round is higher as shown on the below figure, which motivated us to set it as the threshold for the whole training process to get the number of false positive rates.

\begin{figure}
    \centering
    \includegraphics[scale=0.4]{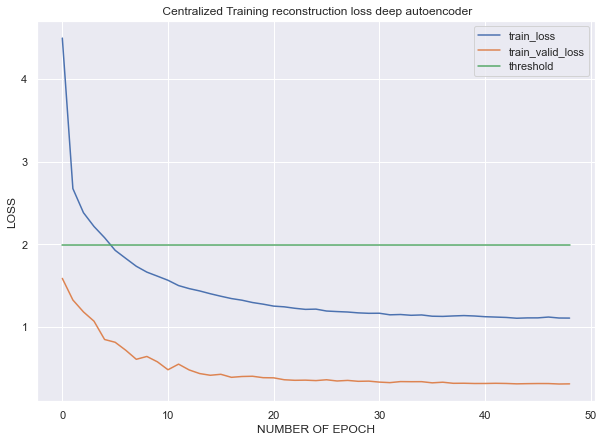}
    \caption{The ML reconstruction loss over 50 epochs}
    \label{fig:train loss}
\end{figure}

\begin{figure}[h]
  \centering
  \subfloat{\includegraphics[width=0.4\textwidth]{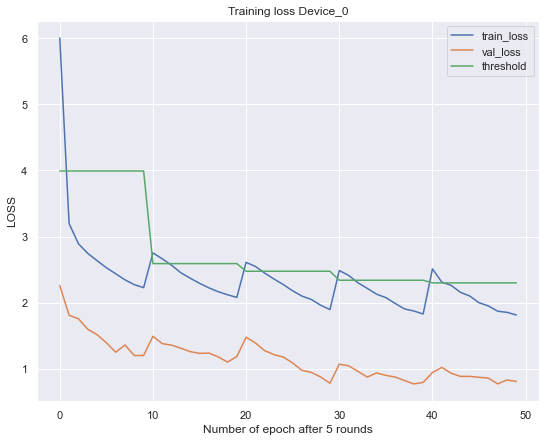}
  \label{fig:f1}}
  \hfill
  \subfloat{\includegraphics[width=0.4\textwidth]{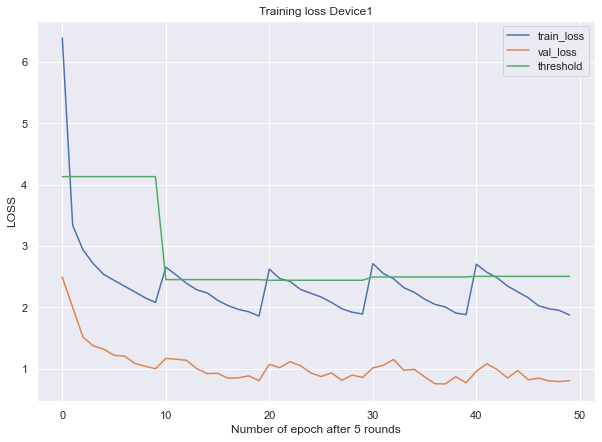}\label{fig:f2}}
  \caption{\emph{Devices with ID 1 and   0 synchronous  FedAvg training reconstruction }}
\end{figure}

\subsection{Testing Results  }
\label{ch:Testing Results}
Testing results for both centralized and decentralized approaches were computed with the help of MSE reconstruction loss and the threshold of the training phase. Testing uses the threshold received from the training to produce our accuracy, precision, recall, and F-Measure scores for the experiment as table II shows.  Test results for with FL settings, the accuracy is an average of five rounds. We have also plotted a confusion matrix for each client/device and for the baseline model. We have shown the percentage of the true positive, negative, false positive and false negative as shown in figure 6.

\begin{figure}[ht]
  \subfloat{\includegraphics[width=0.5\textwidth]{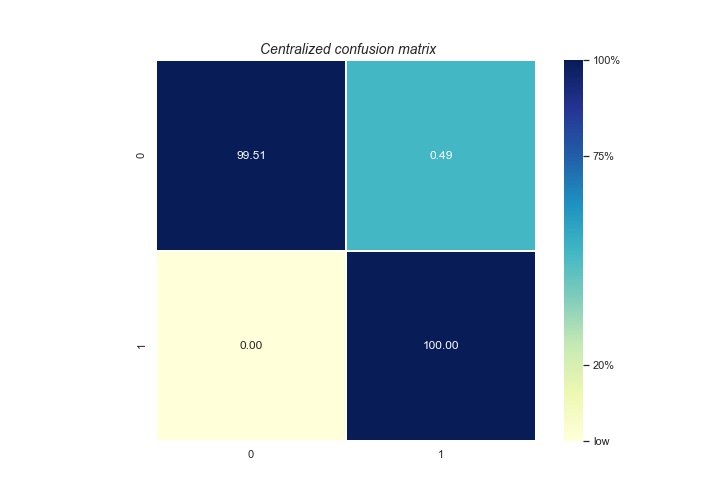}
  \label{fig:matrix_c}}

  \subfloat{\includegraphics[width=0.45\textwidth]{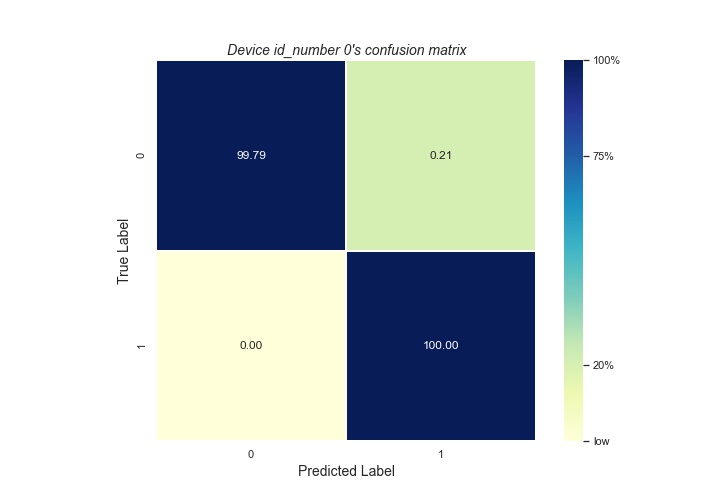}
  \label{fig:matrix0}}
  
  \subfloat{\includegraphics[width=0.45\textwidth]{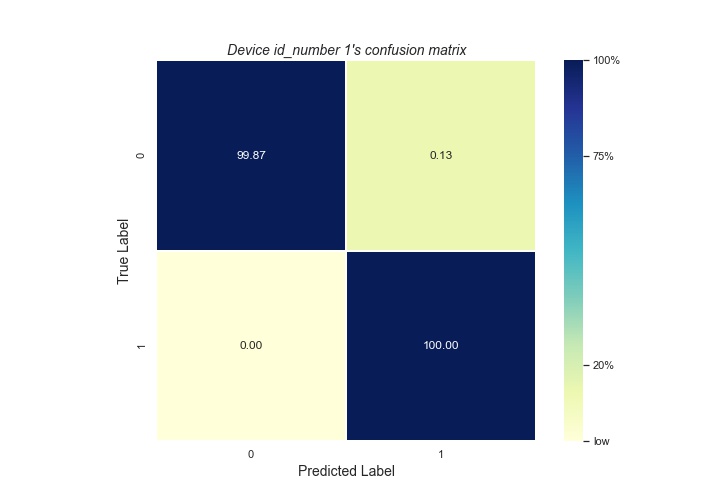}
  \label{fig:matrix1}}
  \caption{\emph{Test confusion  matrix  for both centralize ML and Asynchronous FL }}
\end{figure}

\begin{table}[H] 
    \centering
    \begin{tabular}{|c|c|c|c|}\hline
        &\textbf{Federated}  & &\textbf{Centralized}\\ \hline
         & \textbf{Device 0}& \textbf{Device 1} &   \\\hline
        Accuracy& \textbf{99.8}\% & \textbf{99.8}\%& 99.7\%\\\hline
        Precision & 98.3\% & 98.9\% & \textbf{99.9}\%\\\hline
        Recall & \textbf{100}\% & \textbf{100}\% & \textbf{100}\% \\\hline
        F\_measure& 99.1\%& 99.4\% & \textbf{99.9}\%  \\\hline
        False\_rate &0.2\% & 0.1 \% & \textbf{0.4}\%  \\\hline
    \end{tabular}
    \caption{The comparison of testing results of the ML and Vanilla FL(FedAvg) per each client }
    \label{tab:accuracy_by device}
\end{table}

 
\begin{table}[H] 
    \centering
    \begin{tabular}{|c|c|c|}\hline
        & \textbf{Federated}  &\textbf{Centralized}\\ \hline
        Accuracy& \textbf{99.8}\% & 99.7\%\\\hline
        Precision & 98.6\% & \textbf{99.9}  \% \\\hline
        Recall & \textbf{100}\% & \textbf{100}\% \\\hline
        F\_measure& 99.2\%& \textbf{99.9}\%    \\\hline
        False\_rate & \textbf{0.16\%}  & 0.49\% \\\hline

    \end{tabular}
    \caption{comparison of testing results of the ML and Vanilla FL(FedAvg)   }
    \label{tab:accuracy}
\end{table}

\subsection{Discussion }
\label{ch:Discussion}
Observing the results, the inference is that it is possible for the development of a model for securing modern edge devices in a privacy preserving manner.  Table II and III give the average accuracy of the attack detection using the Vanilla FL approach versus the non-FL baseline. As it can be seen from the figure 7, the accuracy for the federated approach for all experiments came at 99.8\% where the average accuracy of the non-FL baseline reached 99.7\%. Looking at the testing results, the difference seems marginal for the simulation, and the number of false positive rates is better for synchronous FL with 0.16\%. The potential of the federated-based approach could be the effect of training on decentralized data generated from various devices. It remains to be seen if the non-synchronous FL approach, which is our next activity, will also produce better results.

\begin{figure}[h]
    \centering
    \includegraphics[scale=0.4]{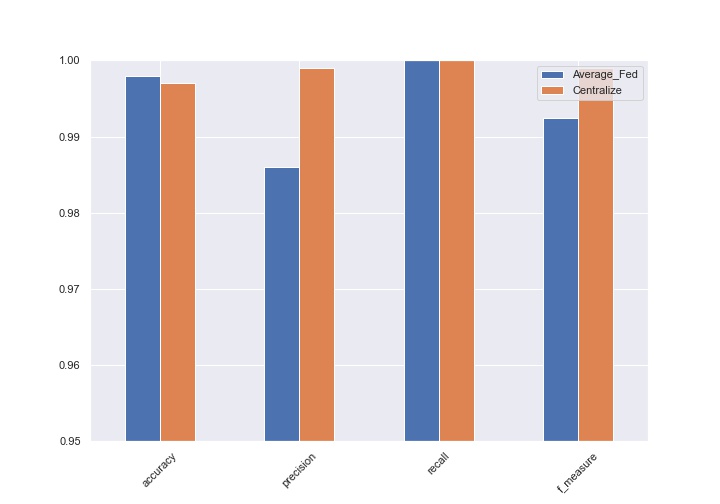}
    \caption{Centralized vs Synchronous FL overall overall-metrics}
    \label{fig:overall_metrics}
\end{figure}


.\section{Conclusions and Future Work}
\label{ch:con}
\subsection{Conclusions}
The ever-growing industries and enterprises usage and diversification of the modern IoT/IIOT devices on their network has posed the risks of new type network attacks. Several approaches have been used to secure them; however, the data protection policy has become a barrier for applying a centralized approach. As such, the usage of Federated learning has been proven to be a solid solution for enterprise IoT/IIoT networks security. It addresses the issue of sensitive data flowing to and from IoT devices being stolen by preserving their privacy when training model is done without sharing of network logs between.

With the online streaming of those devices’ data, labeling data is almost impossible, and they have heterogeneous computing capabilities. For this reason, we applied an autoencoder-based unsupervised approach, and training and evaluating the model was done with the help of federated machine learning. We also designed a fair federated learning algorithm. 

\subsection{Future work}

As shown in Algorithm 1, for FedAvg, at each global iteration,a subset of the devices are selected to run gradient descent optimization (e.g., SGD) locally to  optimize the local objective function \( F_k\) on device k. Then, these local devices communicate their local model updates to the server for aggregation. With heterogeneous local objectives  \( F_k\), carefully tuning of local epochs is crucial for FedAvg to converge. However, a larger number of local epochs may lead each device towards the optima of its local objective as opposed to the central objective. Besides, data continue to be generated on local devices which increases local gradient variations relative to the central model. Therefore, for the future work, we will incorporate a fairness approach to restrict the amount of local deviation by penalizing large changes from the current model at the server. We have  explained this in detail in Section \ref{Federated Averaging}. Moreover, we are considering the issue of computation disparities among the devices participating for collaboratively training the global model. Because of that,the next step is to continue developing and evaluating a Fair federated machine learning using the deep autoencoder with the same settings as centralized and vanilla FedAvg and perform comparisons

\subsection{Limitations} 
Among the activities that were planned, comparing  Fair federated averaging algorithm performance on the new dataset and the vanilla Federated algorithm was on the list. However, due to time constraint of the project, we were not able to fully implement the designed Fair federated averaging algorithm. It is under implementation and considered as future work.




\vspace{12pt}


\end{document}